# Evolving Boolean Regulatory Networks with Epigenetic Control

Larry Bull

Department of Computer Science & Creative Technologies

University of the West of England

Bristol BS16 1QY, U.K.

+44 (0)117 3283161

Larry.Bull@uwe.ac.uk

Abstract

The significant role of epigenetic mechanisms within natural systems has become increasingly clear. This paper uses a recently presented abstract, tunable Boolean genetic regulatory network model to explore aspects of epigenetics. It is shown how dynamically controlling transcription via a DNA methylation-inspired mechanism can be selected for by simulated evolution under various single and multiple cell scenarios. Further, it is shown that the effects of such control can be inherited without detriment to fitness.

1. Introduction

Epigenetics refers to cellular mechanisms that affect transcription without altering DNA sequences, e.g., see [Bird, 2007] for an overview. The two principle mechanisms are DNA methylation and histone modification. In the former case, a methyl group attaches to the base cytosine, or adenine in bacteria, typically causing a reduction in transcription activity in the area. In the latter case, changes in the shape of the proteins around which DNA wraps itself to form chromatin can alter the level of transcription in the area. In both cases, the change can be inherited.

With the aim of enabling the systematic exploration of artificial genetic regulatory network models (GRN), a simple approach to combining them with abstract fitness landscapes has recently been presented [Bull, 2012]. More specifically, random Boolean networks (RBN) [Kauffman, 1969] were combined with the NK model of fitness landscapes [Kauffman & Levin, 1987]. In the combined form – termed the RBNK model – a simple relationship between the states of $N$ randomly assigned nodes within an RBN is assumed such that their value is used within a given NK fitness landscape of trait dependencies. The approach was also extended to enable consideration of multi-celled scenarios using the related NKCS landscapes [Kaufmann & Johnsen, 1991] – termed the RBNKCS model.

In this paper, RBN are extended to include a simple form of epigenetic control. The selective advantage of the new mechanism is explored under various single and multiple celled scenarios. Results indicate epigenetics is useful across a wide range of conditions. The paper is arranged are follows: the next section briefly reviews related work in the area and introduces the two basic models; section 3 examines the extended RBNK model; and, section 4 examines the extended RBNKCS model. Finally, all findings are discussed.

2. Background

2.1 Epigenetic Computing

Whilst there is a growing body of work using artificial GRN within bio-inspired computing (e.g., see [Bull, 2012] for an overview), there are very few examples which consider epigenetic mechanisms explicitly. Note this is not the same as epigenetic robotics (e.g., see [Asada et al., 2009]). Tanev and Yuta [2003] included a histone modification-inspired scheme into a two-cell, rule-based representation where a development phase repeatedly alters one of the two cells. Periyasamy et al. [2008] presented an approach in which each individual in the evolving population is essentially viewed as a protein interacting with other proteins based upon various external and internal conditions, an architecture reminiscent of the Learning Classifier System [Holland, 1976]. Turner et al. [2013] have recently augmented a GRN model with an epigenetic layer in the form of a set of binary masks over the genes, one mask per objective faced by the system: for a given task, the subset of genes defined in the corresponding mask are used to build the GRN. In this paper, a simple, abstract epigenetic mechanism is introduced which is an on-going, context dependent control process during the cell lifecycle.

It can also be noted that a growing number of models of more specific epigenetic mechanisms in organisms exist (e.g., see [Artyomov et al., 2010][Geoghegan & Spencer, 2012][Gupta et al., 2013]).

## 2.2 The RBNK Model

Within the traditional form of RBN, a network of $R$ nodes, each with a randomly assigned Boolean update function and $B$ directed connections randomly assigned from other nodes in the network, all update synchronously based upon the current state of those $B$ nodes. Hence those $B$ nodes are seen to have a regulatory effect upon the given node, specified by the given Boolean function attributed to it. Since they have a finite number of possible states and they are deterministic, such networks eventually fall into an attractor. It is well-established that the value of $B$ affects the emergent behaviour of RBN wherein attractors typically contain an increasing number of states with increasing $B$ (see [Kauffman, 1993] for an overview). Three phases of behaviour exist: ordered when $B=1$, with attractors consisting of one or a few states; chaotic when $B \geq 3$, with a very large number of states per attractor; and, a critical regime around $B=2$, where similar states lie on trajectories that tend to neither diverge nor converge (see [Derrida & Pomeau, 1986] for formal analysis). Figure 1 shows typical behaviour for various $B$.

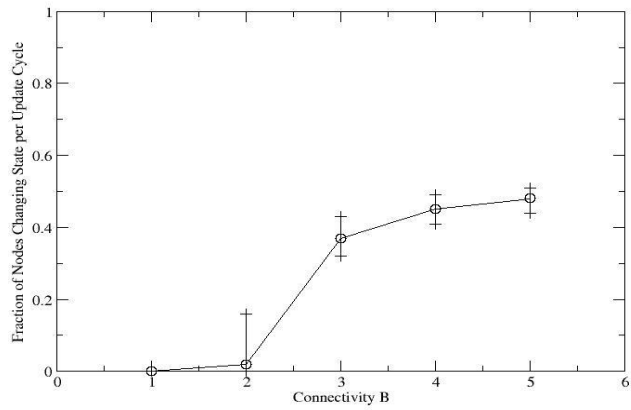

Fig.1. Typical behaviour of RBN with $R=100$ nodes and varying connectivity $B$, averaged after 100 update cycles over 100 runs. Nodes were initialized arbitrarily. Error bars show the min and max behaviour.

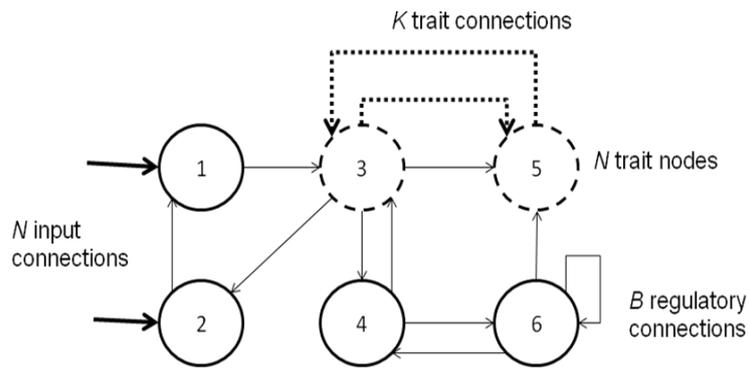

An $R=6$, $B=2$, $N=2$, $K=1$ network with two inputs

Fig. 2. Example RBNK model.

As shown in Figure 2, in the RBNK model $N$ nodes in the RBN are chosen as "outputs", i.e., their state determines fitness using the NK model. Kauffman and Levin [1987] introduced the NK model to allow the systematic study of various aspects of fitness landscapes (see [Kauffman, 1993] for an overview). In the standard NK model an individual is represented by a set of $N$ (binary) genes or traits, each of which depends upon its own value and that of *K* randomly chosen others in the individual. Thus increasing *K*, with respect to *N*, increases the epistasis. This increases the ruggedness of the fitness landscapes by increasing the number of fitness peaks.

The NK model assumes all epistatic interactions are so complex that it is only appropriate to assign (uniform) random values to their effects on fitness. Therefore for each of the possible *K* interactions, a table of $2^{(K+1)}$ fitnesses is created, with all entries in the range 0.0 to 1.0, such that there is one fitness value for each combination of traits. The fitness contribution of each trait is found from its individual table. These fitnesses are then summed and normalised by *N* to give the selective fitness of the individual. Exhaustive search of NK landscapes [Smith & Smith, 1999] suggests three general classes exist: unimodal when *K*=0; uncorrelated, multi-peaked when *K*>3; and, a critical regime around 0<*K*<4, where multiple peaks are correlated.

The combination of the RBN and NK model enables a systematic exploration of the relationship between phenotypic traits and the genetic regulatory network by which they are produced. It was previously shown how achievable fitness decreases with increasing *B*, how increasing *N* with respect to *R* decreases achievable fitness, and how *R* can be decreased without detriment to achievable fitness for low *B* [Bull, 2012]. In this paper *N* phenotypic traits are attributed to arbitrarily chosen nodes within the network of *R* genetic loci, with environmental inputs applied to the first *N'* loci (Figure 2). Hence the NK element creates a tuneable component to the overall fitness landscape with behaviour (potentially) influenced by the environment. For simplicity, *N'*=*N* here.

2.3 The RBNKCS Model

Kauffman highlighted that species do not evolve independently of their ecological partners and subsequently presented a coevolutionary version of the NK model. Here each node/gene is coupled to *K* others locally and to *C* (also randomly chosen) within each of the *S* other species/individuals with which it interacts – the NKCS model [Kauffman & Johnsen, 1991]. Therefore for each of the possible *K*+*C*x*S* interactions, a table of $2^{(K+1+C \times S)}$ fitnesses is created, with all entries in the range 0.0 to 1.0, such that there is one fitness value for each combination of traits. The fitness contribution of each gene is found from its individual table. These fitnesses are then summed and normalised by *N* to give the selective fitness of the total genome (see [Kauffman, 1993] for an overview). It is shown that as *C* increases, mean fitness drops and the time taken to reach an equilibrium point increases, along with an associated decrease in the equilibrium fitness level. That is, adaptive moves made by one partner deform the fitness landscape of its partner(s), with increasing effect for increasing *C*. As in the NK model, it is again assumed all intergenome (*C*) and intragenome (*K*) interactions are so complex that it is only appropriate to assign random values to their effects on fitness.

The RBNK model is easily extended to consider the interaction between multiple GRN based on the NKCS model – the RBNKCS model. As Figure 3 shows, it is here assumed that the current state of the *N* trait nodes of one network provide input to a set of *N* internal nodes in each of its coupled partners, i.e., each serving as one of their *B* connections. Similarly, the fitness contribution of the *N* trait nodes considers not only the *K* local connections but also the *C* connections to its *S* coupled partners' trait nodes. The GRN update alternately.

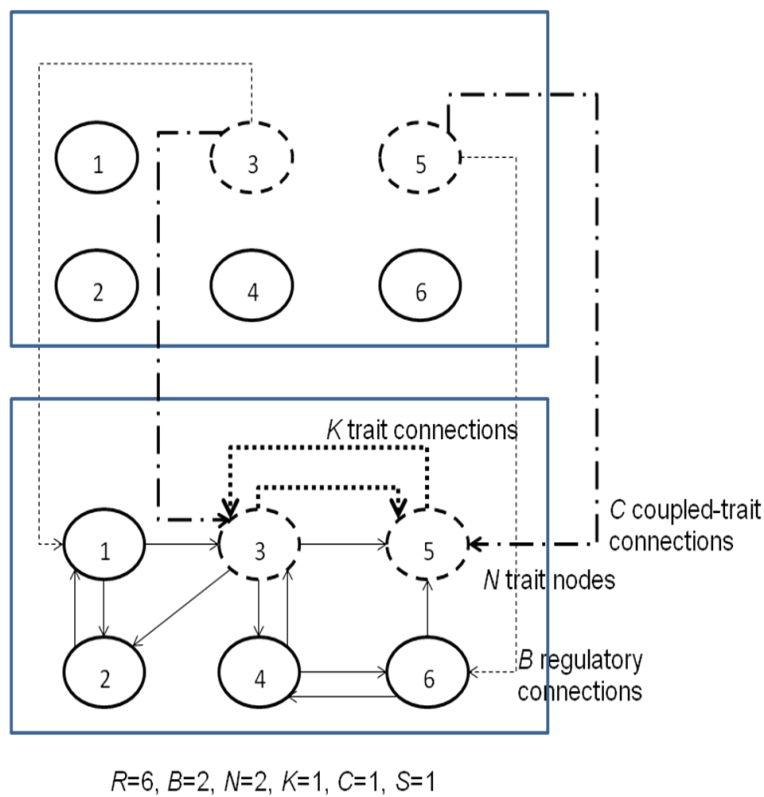

$R=6, B=2, N=2, K=1, C=1, S=1$

Fig. 3. Example RBNKCS model. Connections for only one of the two coupled networks are shown for clarity.

## 3. Epigenetic Control in the RBNK Model

3.1 DNA Methylation

To include a mechanism which enables the suppression of transcription based upon the internal and/or external environment of the cell, each node in the RBN is extended to (potentially) include a second set of $B'$ connections to defined nodes. Each such epigenetic node also performs an assigned control function based upon the current state of the $B'$ nodes, as shown in Figure 4. Hence on each cycle, each unmethylated node updates its state based upon the current state of the $B$ nodes it is connected to using the Boolean logic function assigned to it in the standard way. All nodes also update their methylation status according to the current state of the $B'$ nodes each is connected to. If a node is not currently methylated and its epigenesis look-up table contains a '1' for the current state of the $B'$ network nodes, the node is said to become suppressed due to the attachment of a methyl group and its state set to '0' (off). If a node is currently methylated and its epigenesis look-up table contains a '0' for the current state of the $B'$ network nodes, the node is said to remove the methyl group, with its state left as '0' for that cycle. For simplicity, the number of standard regulatory connections is assumed to be the same as for epigenetic control, i.e., $B=B'$.

3.2 Experimentation

For simplicity with respect to the underlying evolutionary search process, a genetic hill-climber is considered here, as in [Bull, 2012]. Each RBN is represented as a list to define each node's start state, Boolean function, $B$ connection ids, $B'$ connection ids, epigenetic control table entries, and whether it is an epigenetic node or not. Mutation can therefore either (with equal probability): alter the Boolean function of a randomly chosen node; alter a randomly chosen $B$ connection; alter a node start state; turn a node into or out of being an epigenetically controlled node; alter one of the control function entries if it is an epigenetic node; or, alter a randomly chosen $B'$ connection, again only if it is an epigenetic node. A single fitness evaluation of a given GRN is ascertained by updating each node for 100 cycles from the genome defined start states. An input string of $N'$ 0's is applied on every cycle here. At each update cycle, the value of each of the $N$ trait nodes in the GRN is used to calculate fitness on the given NK landscape. The final fitness assigned to the GRN is the average over 100 such updates here. A mutated GRN becomes the parent for the next generation if its fitness is higher than that of the original.

In the case of fitness ties the number of epigenetic nodes is considered, with the smaller number favoured, the decision being arbitrary upon a further tie. Hence there is a slight selective pressure against epigenetic control. Here $R=100$, $N=10$ and results are averaged over 100 runs - 10 runs on each of 10 landscapes per parameter configuration - for 30,000 generations. As in [Bull, 2012], $0<B\leq5$ and $0\leq K\leq5$ are used.

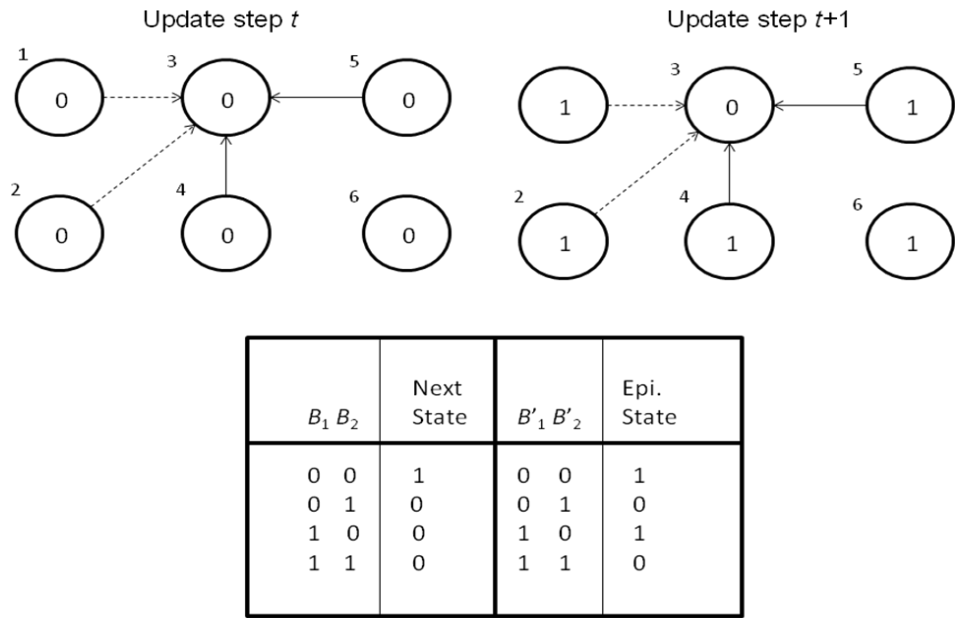

Fig. 4. Example RBN with epigenetic control. The look-up table and connections for node 3 are shown in an $R=6$, $B=2$ network. Nodes susceptible to further control have $B'$ extra regulation connections into the network (dashed arrows) and use the state of those nodes to alter the effects of the standard transcription regulation connections (solid arrows) on the next update cycle ($B'=2$). Thus in the RBN shown, node 3 is an epigenetic node and uses nodes 1 and 2 to determine any higher-level transcription control. At update step $t$, assuming all nodes are at state '0', the given node above would *not* transit to state '1' for the next cycle but would be turned off due to the epigenetic mechanism, as defined in the first row of the table shown. A DNA methylation-like mediated change in the regulation network is said to have occurred and the gene suppressed. Note that at update step $t+2$, the epigenetic control would be relinquished.

As Figure 5 shows, regardless of $K$, epigenetic control is selected for in all high connectivity cases, i.e., when $B>3$. Moreover, the fitness level reached in all such cases is significantly higher than without epigenetic control (T-test, $p<0.05$), i.e., the standard case [Bull, 2012], when the underlying fitness landscape is correlated, i.e., when $K<5$ (not shown). Analysis of the underlying behaviour of the epigenetic nodes indicates that they constantly switch the mechanism on and off during the lifecycle, staying in either state for an unchanging numbers of updates as the RBN falls into an attractor. That is, the epigenetic control has a role in the active part of the resulting cyclic behaviour of the network; a sub-set of nodes are not simply switched off. The period of existing in either state typically ranges from 2-5 update cycles. This general result was also found for other values of *R*, e.g., *R*=200 (not shown).

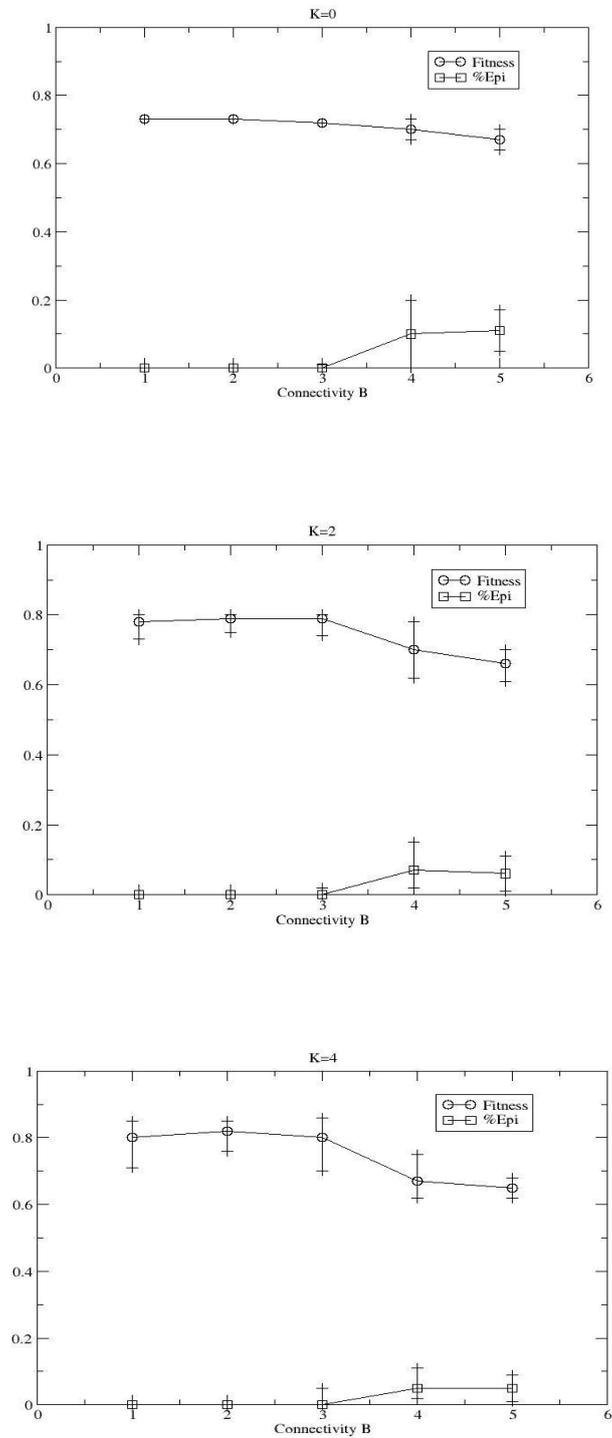

Fig. 5. Evolutionary performance of the epigenetically controlled RBN in various environments, after 30,000 generations. Error bars show the min and max behaviour.

It has long been known (e.g., see [Kauffman, 1984]) that a bias in the transition space of traditional RBN, i.e., away from the expected average of 0.5 for either state, reduces the number of attractors and their size for a given size and connectivity. Thus whilst the evolutionary process (used here) typically struggles to search the node function space to exploit this fact and thereby potentially mitigate some of the effects of high $B$, the epigenetic mechanism appears to provide an easier route to effective control of the underlying chaotic dynamics of such RBN. It can be noted that within natural GRN genes have low connectivity on average (e.g., see [Leclerc, 2008]), as RBN predict, but a number of high connectivity "hub" genes perform significant roles (e.g., see [Barabási & Oltvai, 2004]). The results here suggest epigenetic control may help facilitate such structures.

As mentioned above, epigenetic modifications can be inherited (e.g., see [Franklin & Mansuy, 2010]). In the above experiments, all life-time induced epigenetic control was assumed to be reset during meiosis. Figure 6 shows example results in which the epigenetic status was not reset in the offspring. In all cases it was found that whilst there is no significant change in the fitness level reached (T-test, $p \geq 0.05$) there is typically a significant (T-test, $p<0.05$) increase in the number of epigenetic nodes for $B>3$, regardless of $K$. That is, the evolutionary process appears able to find ways to further exploit epigenetic control over highly connected genes when it is heritable. This was found to be the case regardless of whether node states were also reset or not (not shown).

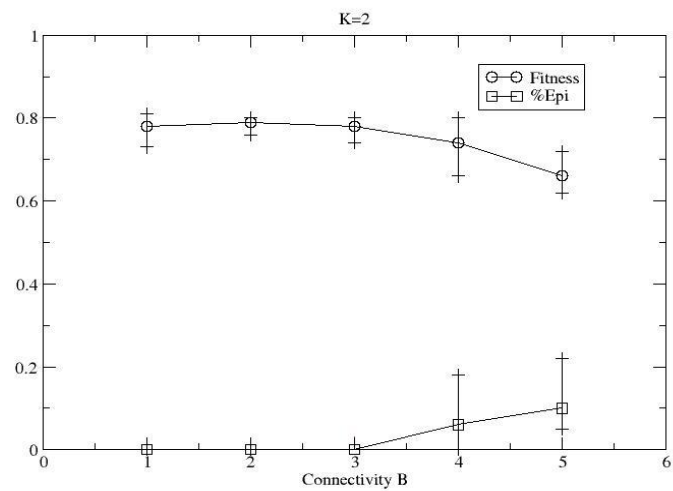

Fig. 6. Example evolutionary performance of the epigenetically controlled RBN, after 30,000 generations, when the mechanism's effects are transgenerational. Error bars show the min and max behaviour.

Moreover, as reviewed in [Franklin & Mansuy, 2010] for example, environmental effects such as toxins or significant changes in nutrients can induce transgenerational epigenetic modifications. This can be explored within the RBNK model through the use of multiple landscapes and corresponding environmental stimuli. Figure 7 shows example results from when the above variant with the inheritance of epigenetic status switches between two NK fitness landscapes A and B. The initial parent RBN experienced an input of $N'$ 0's for the first half of its life with its fitness calculated on landscape A. For the second half of its life, it experienced an input of all 1's with its fitness calculated on landscape B. Its mutated offspring then experienced the reverse scenario, i.e., starting on landscape B before switching to A. Once the parent is replaced by an offspring, as described above, the scenario switches back, etc. In this way an offspring initially exists within the environment last experienced by its parent but which subsequently changes significantly, e.g., due to an increase/decrease in nutrient levels. Figure 7 shows how epigenetic control is now selected for under all conditions, that is for *all* values of *B*, without significant variance in the number of epigenetic nodes or fitness (T-test, $p \geq 0.05$), regardless of *K*. Analysis of the underlying behaviour shows different nodes becoming epigenetically controlled with the switch in the input stimulus.

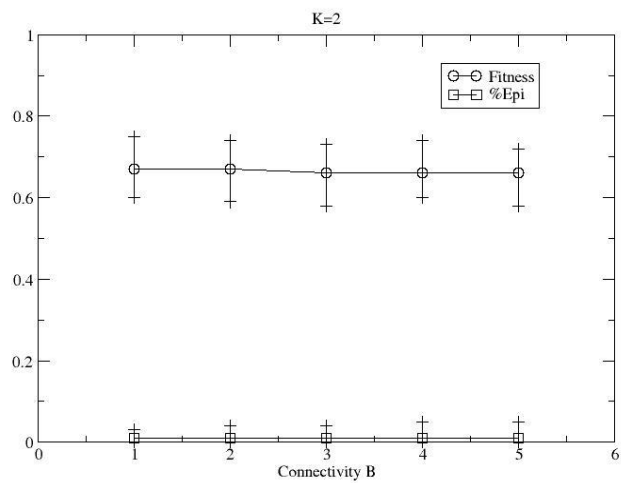

Fig. 7. Example evolutionary performance of the transgenerational epigenetically controlled RBN in a changing environment, after 30,000 generations.

## 4. Epigenetic Control in the RBNKCS Model

The role of epigenetic control in the development of multicellular organisms is clear (e.g., see [Kieffer, 2007]). The simple case of two interacting cells has previously been explored with the RBNKCS model, as shown in Figure 3, where one is the daughter (genetic clone) of the other, i.e., $S=1$ [Bull, 2012] (after [Bull, 1999]). Figure 8 shows example results for various $B$, $K$, and $C$ where the mother and daughter cells exist on two different NKCS landscapes, with fitness averaged. That is, some functional differentiation in the phenotype of the cells is anticipated. All other details were as above and $0<C\leq 5$. See [Bull, 2012] for an overview of previous work using coupled GRN.

The results indicate that for $B=1$, regardless of $K$ and $C$, epigenetic control is not selected for. Recall such networks typically enter a point attractor quickly (Figure 1) and hence the GRN of the two cells exist in relatively static environments. Unlike in the single-celled case above (Figure 5), for all $B>1$ epigenetic control is selected for in around 5-10% of nodes, over all $K$ and $C$. Again, in comparison to the equivalent traditional scenario without epigenetics [Bull, 2012], for low $C$, the fitness level reached in all such cases is significantly higher (T-test, $p<0.05$) with epigenetic control (not shown), except for $B=2$, where the difference is not consistently significantly different. That is, evolution appears to exploit epigenetic control to mutually shape the attractors of the two interacting cells such that a high fitness level can be reached reliably: cooperative behaviour between the two dynamical systems is enhanced by the extra layer of control. For high $C$, e.g., $C=5$, there is no significant benefit seen from epigenetic control over the traditional case but it is still typically selected for when $B>1$. It can be noted that results from the homogeneous case wherein the mother and daughter cells exist on the same NKCS landscape are equivalent to those seen above in the single cell case in terms of both the conditions and degree of adoption of epigenetic control, i.e., when $B>3$ and around 5% of nodes (not shown).

The multicellular variant of the RBNKCS used previously and above does not include a period of something akin to *in utero* development. That is, each GRN and its cloned copy is set to its start state and executes in turn, with the fitness contribution of the $N$ external traits recorded per cycle. Given the role of epigenetics in development, this basic model has been extended to include a period of execution during which fitness is not calculated. Here each genome includes an integer $D$ defining how many update cycles to execute before the 100 fitness calculating updates as above. This value is here initialized to zero and the mutation operator extended to randomly increment or decrement it. The replacement process is modified such that, if fitness and the number of epigenetic nodes are both equal, the smaller *D* is chosen, otherwise the decision is random.

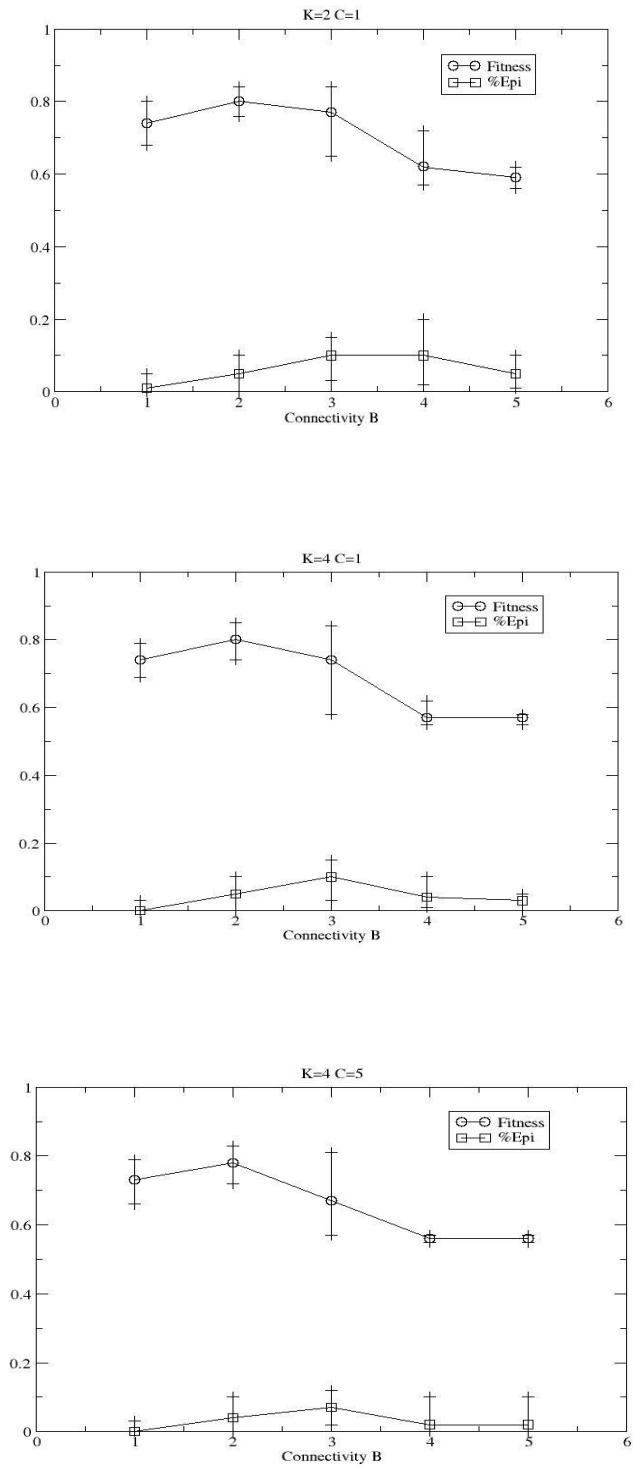

Fig. 8. Evolutionary performance of the epigenetically controlled RBN in a multicelled environment, after 30,000 generations. Error bars show the min and max behaviour.

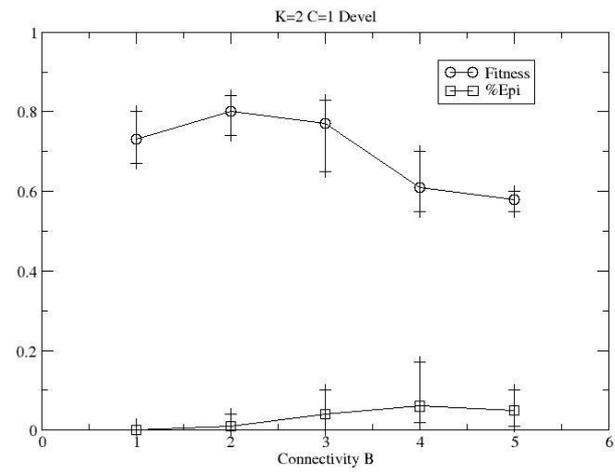

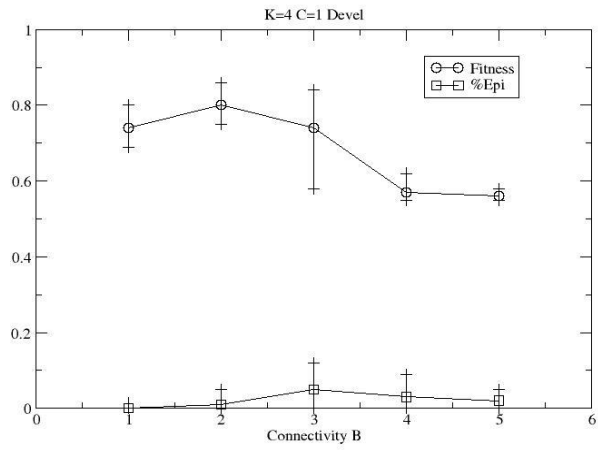

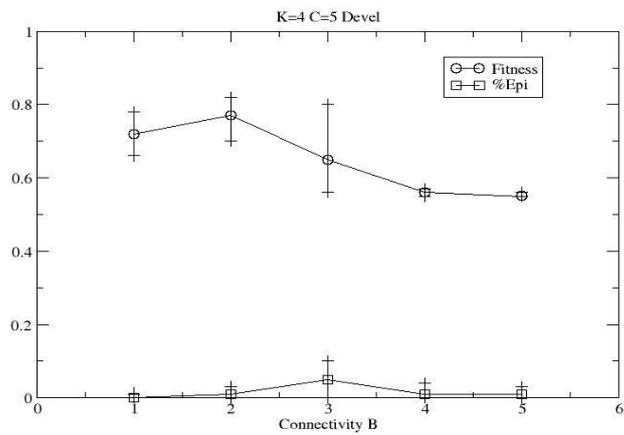

Fig. 9. Evolutionary performance of the epigenetically controlled RBN in a multicelled environment with a development phase added, after 30,000 generations. Error bars show the min and max behaviour.

As Figure 9 shows, the added development phase typically reduces the fraction of nodes using epigenetic control, usually but not always with statistical significance (T-test, p<0.05), especially for 1<B<4 and low C. Figure 10 shows the average values of D for the cases shown in Figure 9. As can be seen, the phase is used most in the cases where the amount of epigenetic control is most reduced.

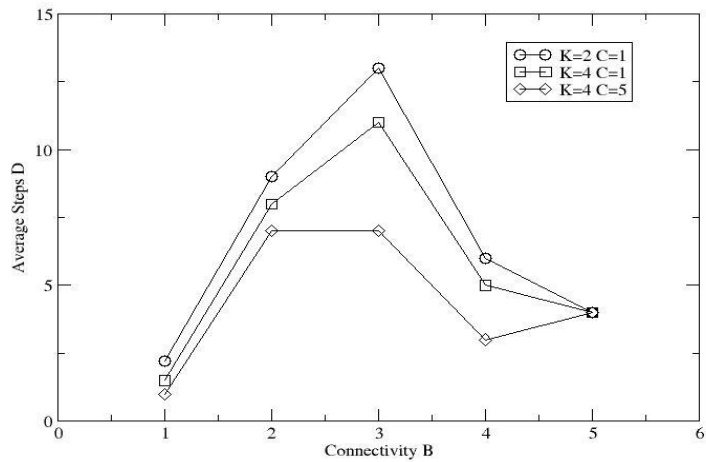

Fig. 10. Number of pre-fitness calculation development steps which evolved for the cases shown in Figure 9.

Finally, the removal of resetting epigenetic controls under meiosis has been explored here, as in Figure 6, and results indicate no significant change in fitness, the number of nodes using epigenetic control, nor in the number of development steps used (not shown).

5. Conclusions

There is a growing body of work exploring artificial genetic regulatory networks, both as a representation scheme for machine learning and as a tool for systems biology. In particular, adoption of these relative generic representations creates the opportunity to exploit new mechanisms more closely based upon the biology. This paper has explored the use of epigenetic control via a DNA methylation-inspired scheme within a well-known abstract GRN model. It has been shown that simple epigenetic control is positively selected for in highly connected GRN in all environments considered, for all degrees of connectivity within non-stationary environments, and for all but the least connected networks with multicelluar/coupled GRN. Moreover, any epigenetic control occurring during a parent's lifecycle can be inherited by the offspring without detriment.

Further consideration of epigenetic control suggests a number of extensions to this work in the near future, particularly in conjunction with the use of mobile DNA inspired mechanisms (e.g., see [Bull, 2013]). Future work should also consider the use of epigenetic control within other GRN representations, explore whether these results also hold for asynchronous GRN updating schemes, and determine whether the general results also appear to be true for much larger multicellular systems, i.e., for $S>>1$.